\documentclass{ecai}
\usepackage{times}
\usepackage{latexsym}
\usepackage{times}
\usepackage{graphicx}
\usepackage{epsfig}
\usepackage{caption}
\usepackage{dirtytalk}
\usepackage{multicol}
\usepackage{multirow}
\usepackage{tabularx}
\usepackage{subcaption}
\begin{document}

\title{Bagged Boosted Trees for Classification of Ecological Momentary Assessment Data}

\author{Gerasimos Spanakis \and Gerhard Weiss \institute{Department of Data Science and Knowledge Engineering, Maastricht University, email:\{jerry.spanakis,gerhard.weiss\}@maastrichtuniversity.nl} \and Anne Roefs\institute{Faculty of Psychology and Neuroscience, Maastricht University, email: a.roefs@maastrichtuniversity.nl}}

\maketitle

\begin{abstract}
Ecological Momentary Assessment (EMA) data is organized in multiple levels (per-subject, per-day, etc.) and this particular structure should be taken into account in machine learning algorithms used in EMA like decision trees and its variants. We propose a new algorithm called BBT (standing for Bagged Boosted Trees) that is enhanced by a over/under sampling method and can provide better estimates for the conditional class probability function. Experimental results on a real-world dataset show that BBT can benefit EMA data classification and performance.
\end{abstract}

\section{Background \& Motivation}
\label{sect:intro}
This work focuses on classification trees and how their ensembles can be utilized in order to set up a prediction environment using Ecological Momentary Assessment (EMA) data from a real-world study. EMA \cite{shiffman2008ecological} refers to a collection of methods used in many different disciplines by which a research subject repeatedly reports on specific variables measured close in time to experience and in the subject's natural environment (e.g. experiencing food craving is measured again and again on the same subject). EMA aims to minimize recall bias, maximize ecological validity and allow microscopic analysis of influence behavior in real-world contexts. EMA data has a different structure than normal data and account for several dependencies between them, since e.g. many samples belong to the same subject so they are expected to be correlated. However, most decision trees that deal with EMA data do not take these specificities into account.

Bagging involves having each tree in the ensemble vote with equal weight while boosting involves incrementally building an ensemble by training each new model instance to emphasize the training instances that previous models mis-classified. Major differences between bagging and boosting are that (a) boosting changes the distribution of training data based on the performance of classifiers created up to that point (bagging acts stochastically) and (b) bagging uses equal weight voting while boosting uses a function of the performance of a classifier as a weight for voting.

There are limited studies on combining bagging and boosting (\cite{suen2005combining}, \cite{kotsiantis2004combining}, \cite{Quinlan:1993:CPM:152181} and \cite{xie2009combination}), however, none of these approaches have been applied to longitudinal data or take into account the EMA structure. Efforts to apply decision trees to EMA data have been attempted but they are mostly focusing on regression tasks (\cite{sela2012re}, \cite{loh2013regression}, \cite{fu2015unbiased}) and on the other hand they do not use bagging or boosting for improving performance. Work in current paper aims at bridging this gap by combining bagging and boosting with the longitudinal data structure. 

\section{BBT: The proposed algorithm}
\label{sect:algo}
Let the training data be $x_1,...,x_n$ and $y_1,...,y_n$ where each $x_i$ is a $d$-dimensional vector and $y_i \in \{-1,1\}$ is the associated observed class label. To justify generalization, it is usually assumed that training data as well as any test data are \textit{iid} samples from some population of $(x,y)$ pairs. Our goal is to as accurately predict ${y}_{i}$ given $x_{i}$. 

The first step to fit a BBT is to select the loss function, which in the case of a classification problem is based on the logistic regression loss. After some initial parameter selection (number of trees to be grown in sequence, shrinkage (or learning) rate, size of individual trees and fraction of the training data sampled) we grow BBT (say using $M$ trees) on the training data using the following process and by growing single Boosted Trees (BT):
\begin{itemize}
\item Divide the data into $B$ (typically $5-10$) subsets and construct $B$ training data sets each of which omits one of the $B$ subsets (the \lq out-of-bag\rq\ data). Each one of the $B$ subsets is created by bootstrap sampling data points from the set of subjects ($p=1,...,P$). To create the learning set we introduce the strategy $\mathcal{S}$ according to which one observation is drawn per subject. This strategy is based on a simple rationale: When only one observation per subject is selected, the probability that different observations are used for the training of different trees is increased, although the same subjects might be selected which further reduces similarity between trees. By this way, we manage to incorporate advantages of subject-based bootstrapping and observation-based bootstrapping into the final BBT ensemble. Also, this approach can be applied to unbalanced data points per subject.
\item Grow $B$ BT; one for each of the $B$ training sets, based on the AdaBoost algorithm \cite{freund1996experiments}: First let $F_0(x_i)=0$ for all $x_i$ and initialize weights $w_i = 1/d$ for $i=1,...,d$. Then repeat the following for $m=1,...,M$ for each one of the $B$ BT:
\begin{itemize}
\item Fit the decision tree $g_m$ to the training data sample using weights $w_i$ where $g_m$ maps each $x_i$ to -1 or 1.
\item Compute:\\
- the weighted error rate $\epsilon_m = \sum_{i=1}^n w_iI\{y_i \neq g_m(x_i)\}$\\
- half its log-odds and derive $\alpha_m = \frac{1}{2} \log\frac{1-\epsilon_m}{\epsilon_m}$
\item Let $F_m = F_{m-1} + \alpha_mg_m$ 
\item Replace the weights $w_i$ with $w_i = w_i e^{-\alpha_mg_m(x_i)y_i}$ and then renormalize by replacing each $wi$ by $wi/(\sum w_i)$.
\end{itemize}
\item Calculate the PE for each BT for tree sizes $1$ to $M$ from the corresponding out-of-bag data and pool across the $B$ boosted trees. Predictions for new data are computed by first predicting each of the component trees and then aggregate the predictions (e.g., by averaging), like in bagging.
\item The minimum PE estimates the optimum number of trees $m*$ for the BT. The estimated PE of the single BT obtained by cross-validation can thus also be used to estimate PE for the BBT. BBT thus require minimal additional computation beyond estimation of $m*$.
\item Reduce the number of trees for each BT to $m*$.
\end{itemize}

For a classification problem, we use an estimate $p_m(x)$ of the Conditional Class Probability Function (CCPF) $p(x)$ that can be obtained from $F_m$ through a logistic link function:

\begin{equation}
p_m(x) = p_m(y = 1|x) = \frac{1}{1+\exp({-2F_m(x))}}
\end{equation}

Classifying at the 1/2 quantile of the CCPF works well for binary classification problems but in the case of EMA data, sometimes classification with unequal costs or, equivalently, classification at quantiles other than 1/2 is needed. Strategies about correctly computing the CCPF are considered \cite{Mease2007} by over/under-sampling which convert a median classifier into a q-classifier.

\section{Experiments}
\label{sect:exp}
\begin{figure}[!h]
\centering
\scalebox{0.95}{
  \includegraphics[width=1.0\linewidth]{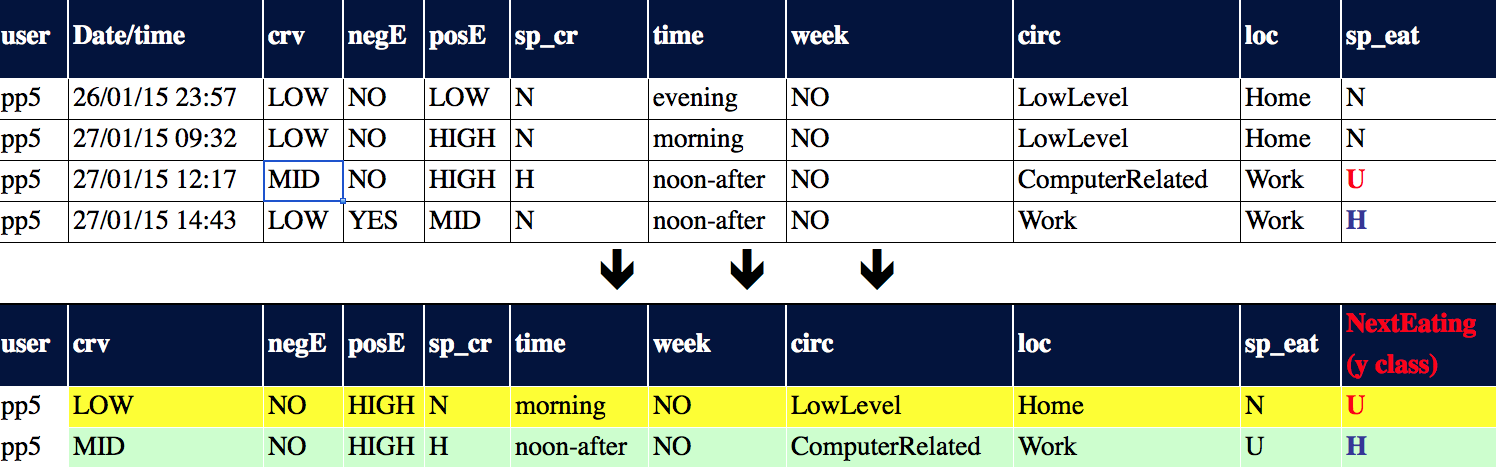}}
    \captionsetup{justification=centering}
    \caption{Data conversion example for early prediction}
  \label{fig:data}
\end{figure}

In order to illustrate the effect of BBT, we now apply this method to an EMA dataset obtained by a study designed by the authors \cite{spanakis2015}. The EMA study followed 100 participants over the course of 14 days using experience \& event sampling questionnaires ending up with over 5000 data points containing information about participants' eating events, emotions, circumstances, locations, etc. (in total there are 9 variables) for several time moments during each day that they participated in the study. Each data point is used to predict whether the next data point (provided that they both occur on the same day) will be a healthy or an unhealthy eating moment. Figure \ref{fig:data} shows an example of how data points (belonging to user \say{pp5}) are converted and combined in order to enable early prediction using a classification algorithm (class can be either \say{healthy} or \say{unhealthy}). Then the BBT algorithm can be applied.

In the comparison between methods, BBT gave a PE of $23.3\%$, whereas the single classification tree ($37.3\%$), bagged trees ($28.9\%$), boosted trees (adaboost) ($25.9\%$), random forests ($26.8\%$) and B\&B combine method \cite{xie2009combination} ($26.2\%$) have higher PE than the BBT. Table \ref{tab:results} summarizes these results and also presents a series of experiments made to demonstrate the effectiveness of  BBT when the number of different subjects ($P$) involved in the dataset increases. For relatively small numbers of subjects (10 or 20) performance of BBT and AdaBoost is comparable (although variance increases and the number of data samples is not large enough) but as $P$ increases the performance of BBT is clearly better. Larger $P$ means that there are more subjects in the dataset, thus the complexity of longitudinal structure increases and it is  imperative to take this into account when classifying longitudinal data. This is the reason that BBT performs better than all other algorithms as $P$ increases. However, for small $P$ the effect of different subjects is smaller and this is the reason that Adaboost performs slightly better than all other algorithms. 

\begin{table}
  \centering
  \captionsetup{justification=centering}
  \caption{Prediction Error (Variance) \% for different algorithms and different numbers of subjects ($P$)}
    \begin{tabular}{rcccc}
    \hline
          & {P=10} & {P=20} & {P=50} & {P=100} \\
    \hline
    \multicolumn{1}{c}{{SCT}} & 25.1 (0.10) & 27.4 (0.08) & 30.9 (0.10) & 37.3 (0.06) \\
    \multicolumn{1}{c}{{Bagging}} & 24.4 (0.22) & 23.8 (0.08) & 30.1 (0.08) & 28.9 (0.06) \\
    \multicolumn{1}{c}{{Boosting}} & \textbf{22.0 (0.12)} & 22.7 (0.10) & 27.0 (0.06) & 25.9 (0.04) \\
    \multicolumn{1}{c}{{Random Forest}} & 23.7 (0.16) & 24.5 (0.14) & 27.2 (0.04) & 26.8 (0.04) \\
    \multicolumn{1}{c}{{B\&B} Combine} & 23.2 (0.08) & 25.1 (0.06) & 26.8 (0.08) & 26.2 (0.02) \\
    \multicolumn{1}{c}{{BBT}} & 22.4 (0.14) & \textbf{21.9 (0.06)} & \textbf{24.2 (0.04)} & \textbf{23.3 (0.02)} \\
    \hline
    \end{tabular}%
  \label{tab:results}
  \end{table}

%
%
%

\section{Discussion}
\label{sect:concl}
In this paper a combination of bagging and boosting was presented: Bagged Boosted Trees (BBT). BBT have the advantage of being able to deal with multiple categorical data which raises a scalability issue when dealing with classic models (like generalized linear models) that are widely used in EMA studies. Moreover, BBT can tackle potential nonlinearities and interactions in the data, since these issues are handled through the combination of many different trees of different sizes. Experimental results of BBT on a real-world EMA dataset clearly show improvement with respect to accuracy in prediction compared to other decision tree algorithms. Further work involves the evaluation of the conditional class probability function (based on over/under sampling of data), as well as the application to other EMA datasets. Finally, adjustment of boosting in order to implement weights based on subjects (and not individual observations) is a direction with promising results.


\bibliographystyle{ecai}

\end{document}